# On Constructing the Value Function for Optimal Trajectory Problem and its Application to Image Processing

**Myong-Song HO**, **Gwang-Hui JU**, **Yong-Bom O** and **Gwang-Ho JONG**

Faculty of Mathematics, **Kim Il Sung** University, D. P. R. Korea

## Abstract

We proposed an algorithm for solving Hamilton-Jacobi equation associated to an optimal trajectory problem for a vehicle moving inside the pre-specified domain with the speed depending upon the direction of the motion and current position of the vehicle. The dynamics of the vehicle is defined by an ordinary differential equation, the right hand of which is given by product of control(a time dependent fuction) and a function dependent on trajectory and control. At some unspecified terminal time, the vehicle reaches the boundary of the pre-specified domain and incurs a terminal cost. We also associate the traveling cost with a type of integral to the trajectory followed by vehicle. We are interested in a numerical method for finding a trajectory that minimizes the sum of the traveling cost and terminal cost. We developed an algorithm solving the value function for general trajectory optimization problem. Our algorithm is closely related to the Tsitsiklis's Fast Marching Method and J. A. Sethian's OUM and SLF-LLL[1-4] and is a generalization of them. On the basis of these results, We applied our algorithm to the image processing such as fingerprint verification.

This paper is concerned with the constructing the value function for an optimal trajectory problem and its application to fingerprint verification.

In literature [1], an algorithm constructing the value function for an optimal exit time control problem is considered and in literature [2-4], algorithms constructing the value function for various type of the min-time control problems are proposed.

I have proved some properties of the value function for general optimal trajectory problem and on the basis of these results, have proposed a new algorithm constructing the value function, and prov-ed its convergence.

Consider an optimal trajectory problem for a vehicle moving inside the domain $\Omega$, with the speed $f$ depending upon the direction of motion and the current position of the vehicle.

The dynamics of the vehicle is defined by $\dfrac{dy(t)}{dt} = f(y(t),\, \alpha(t)) \cdot \alpha(t),\; y(0) = x \in \Omega$, where $y(t)$ is the position of the vehicle at time $t$, $S_1 := \{a \in R^n \mid \|a\| = 1\}$ is the set of admissible control values, and $\Lambda := \{\alpha : [0,\, +\infty) \to S_1 \mid \alpha(\cdot) : \text{measurable}\}$ is the set of admissible controls.

I am interested in studying $y(t)$ only while the vehicle remains inside $\Omega$, i.e., until the exit ti-me $t_x(\alpha) := \{t \geq 0 \mid y_{x,\,\alpha}(t) \in \partial\Omega\}$, where $y_{x,\,\alpha}(t)$ is the position of the vehicle at time corresponding initial state $x$ and control $\alpha(\cdot)$.

If the traveling cost corresponding to the trajectory followed by vehicle is $\displaystyle\int_0^{t_x(\alpha)} l(y_{x,\,\alpha}(t),\, \alpha(t))dt$ and the terminal cost for exiting the domain at the point $x \in \partial\Omega$ is $q(y_{x,\,\alpha}(t_x(\alpha)))$, then an optimal trajectory problem is to find an optimal control $\alpha(\cdot)$ which minimizes the sum of the traveling cost and terminal cost.

The value function of the optimal trajectory problem is as follows.

$$\upsilon(x) = \begin{cases} \displaystyle\inf_{\alpha(\cdot) \in \Lambda} J(x,\, \alpha), & x \in \Omega \\ q(x), & x \in \partial\Omega \end{cases},$$

where $J(x,\, \alpha) = \displaystyle\int_0^{t_x(\alpha)} l(y_{x,\,\alpha}(t),\, \alpha(t))dt + q(y_{x,\,\alpha}(t_x(\alpha)))$.

I will assume that $f$, $l$ and $q$ are Lipschitz-continuous and that there exist constants $f_1,\, f_2,\, l_1,\, l_2,\, q_1,\, q_2$ such that

$$\left.\begin{aligned} 0 < f_1 \leq f(x,\, a) \leq f_2 < +\infty, &\quad \forall x \in \Omega,\; \forall a \in S_1 \\ 0 < l_1 \leq l(x,\, a) \leq l_2 < +\infty, &\quad \forall x \in \Omega,\; \forall a \in S_1 \\ 0 < q_1 \leq q(x) \leq q_2 < +\infty, &\quad \forall x \in \partial\Omega \end{aligned}\right\}.$$

### 1. Properties of the value function

**Lemma 1** Let $d(x)$ be the minimum distance to the boundary $\partial\Omega$. Then for every point $x \in \Omega \setminus \partial\Omega$ and for any $\tau < d(x)$, $\upsilon(x) = \inf\left\{\displaystyle\int_0^\tau l(y_{x,\,\alpha}(s),\, \alpha(s))ds + \upsilon(y_{x,\,\alpha}(\tau))\right\}$.

**Lemma 2** For every point $x \in \Omega$, $l_1 d(x)/f_2 + q_1 \leq \upsilon(x) \leq l_2 d(x)/f_1 + q_2$.

Where $d(x)$ is the distance to the boundary $\partial\Omega$.

**Lemma 3** The value function $\upsilon(x)$ is Lipschitz-continuous on $\Omega \setminus \partial\Omega$.

**Lemma 4** Consider a point $\bar{x} \in \Omega \setminus \partial\Omega$. Then, for any constant $C$ such that $q_2 \leq C \leq \upsilon(\bar{x})$, the optimal trajectory for $\bar{x}$ will intersect the level set $L = \{x \mid \upsilon(x) = C\}$ at some point $\tilde{x}$. If $\bar{x}$ is distance $d_1$ away from that level set $L$, then $\|\tilde{x} - \bar{x}\| \leq d_1 \cdot f_2 l_2 / (f_1 l_1)$.

The Hamilton–Jacobi–Bellman PDE for the value function is as follows.
$$\min_{a \in S_1}\{(Du(x) \cdot a) \cdot f(x, a) / l(x, a)\} + 1 = 0 \qquad (*)$$

**Definition** $u(\cdot)$ is called the viscosity solution of equation (*) if $u(\cdot) \in C(\overline{\Omega})$ satisfies the following conditions.

1) for each function $\varphi \in C^\infty(\Omega)$, if $u - \varphi$ has a local minimum at $x_0 \in \Omega$, then
$$\min_{a \in S_1}\{(D\varphi(x_0) \cdot a) \cdot f(x_0, a) / l(x_0, a)\} + 1 \geq 0.$$

2) for each function $\varphi \in C^\infty(\Omega)$, if $u - \varphi$ has a local maximum at $x_0 \in \Omega$, then
$$\min_{a \in S_1}\{(D\varphi(x_0) \cdot a) \cdot f(x_0, a) / l(x_0, a)\} + 1 \leq 0.$$

## 2. An algorithm constructing the value function

Assume that a triangulated mesh $X$ of diameter $h$ is defined on $\Omega$.

For every mesh point $x \in X$, define $S(x)$ to be a set of all the simplexes in the mesh adjacent to $x$ (i.e., the simplexes that have $x$ as one of their vertices).

If $s \in S(x)$, I will use the notation $x_{s,1}, x_{s,2}$ for the other vertices of the simplex $s$.

I will assume that, as the vehicle starts to move from a mesh point $x$ inside a simplex $s \in S(x)$, its direction $a$ of motion does not change until the vehicle reaches the edge $x_{s,1}, x_{s,2}$.

The value $\upsilon(\tilde{x})$ at the point of intersection can be approximated using the values $\upsilon(x_{s,1})$ and $\upsilon(x_{s,2})$, i.e.
$$\upsilon(\tilde{x}) = \zeta \cdot \upsilon(x_{s,1}) + (1 - \zeta) \cdot \upsilon(x_{s,2}) .\text{(where } \tilde{x} = \zeta \cdot x_{s,1} + (1 - \zeta) \cdot x_{s,2} \ (0 \leq \zeta \leq 1)).$$

Define $d(\zeta) := \|\tilde{x} - x\| = \|\zeta \cdot x_{s,1} + (1 - \zeta) \cdot x_{s,2} - x\|$, $\alpha(t) := a_\zeta = (\tilde{x} - x)/(d(\zeta),$
$$\tau(\zeta) := d(\zeta) / f(x, a_\zeta).$$

I can now write the equation for the numerical approximation $U$:
$$U(x) = \min_{s \in S(x)} V_s(x), \quad V_s(x) := \min_{\zeta \in [0,1]} \{d(\zeta) l(x, a_\zeta) / f(x, a_\zeta) + \zeta \cdot U(x_{s,1}) + (1 - \zeta) \cdot U(x_{s,2})\}.$$

The above derivation is based on a direct application of Bellman's optimality principle rather th-an on discretization of the corresponding Hamilton–Jacobi–Bellman PDE.

All the mesh points are divided into three classes: $S_F$ (the set of mesh points for which no infor-mation about the correct value of $U$ is known), $S_A$ (the set of mesh points for which the correct value of $U$ has been computed), and $S_C$ (the set of mesh points adjacent to $S_A$), for which $V$ has already been computed, but it is still not clear if $V = U$ or not.

For every $x \in S_C$, I define the set
$$NF(x) := \{x_j x_k \in AF \mid \exists \tilde{x} \in L(x_j x_k) : \|\tilde{x} - x\| \leq hf_2 l_2 /(f_1 l_1)\},$$
where
$$L(x_j x_k) := \{x \mid x = \lambda x_j + (1-\lambda) x_k, \ 0 \leq \lambda \leq 1\}.$$

An algorithm constructing the value function is as follows.

**Step 1** Start with putting $S_A = S_C = \phi$ and $S_F = X$ and $V(x) = +\infty$ for all the mesh points $x \in X$.

**Step 2** Move the mesh points $x \in \partial \Omega$ on the boundary to $S_A$ ($V(x) = q(x)$).

**Step 3** Move all the mesh points $x$ adjacent to the boundary into $S_C$ and evaluate the tentative values $U(x) = \min_{s \in NS(x)} V_s(x)$.

**Step 4** Apply SLF-LLL method to find the mesh point $\bar{x}$ with $S_C$.

**Step 5** Move $\bar{x}$ to $S_A$ ($U(\bar{x}) := V(\bar{x})$).

**Step 6** Move the mesh points of $S_F$ adjacent to $\bar{x}$ into $S_C$.

**Step 7** Reevaluate $V$ for all the $x \in S_C$ adjacent to $\bar{x}$ by
$$V(x) := \min\{V(x), \min_{\bar{x} x_i \in NF(x)} V_{\bar{x} x_i}(x)\}.$$

**Step 8** If $S_C$ is not empty, then go to 4.

By using the above algorithm I have the numerical solution $U(x)$ of the value function on each mesh point $x \in X$.

For the points $x \in \overline{\Omega} \setminus X$, I can define $U(x)$ by the following linear interpolation.

Let us assume that $x \in \overline{\Omega} \setminus X$ lies in some simplex $x_1, x_2, x_3$.

Where $x_1, x_2, x_3$ are mesh point.

In that case, there exist $\zeta_1, \zeta_2, \zeta_3 \geq 0$ such that
$$\sum_{i=1}^{3} \zeta_i = 1, \ x = \sum_{i=1}^{3} \zeta_i x_i.$$
The value $U(x)$ at $x \in \overline{\Omega}$ is defined to be
$$U(x) := \zeta_1 U(x_1) + \zeta_2 U(x_2) + \zeta_3 U(x_3).$$

**Theorem** Let $U^r$ be the approximate solution obtained on the mesh $X_r$ by the algorithm descry-bed above. As $h_r \to 0$, $U^r$ uniformly converges to the viscosity solution of the Hamilton–Jacobi–Bellman PDE (*).

### 3. Applying to Image processing

I have applied our algorithm to the image processing for Fingerprint verification.

As the result, I extracted the minutiae (bifurcation points, end points) of fingerprint image with noise as the following figure.

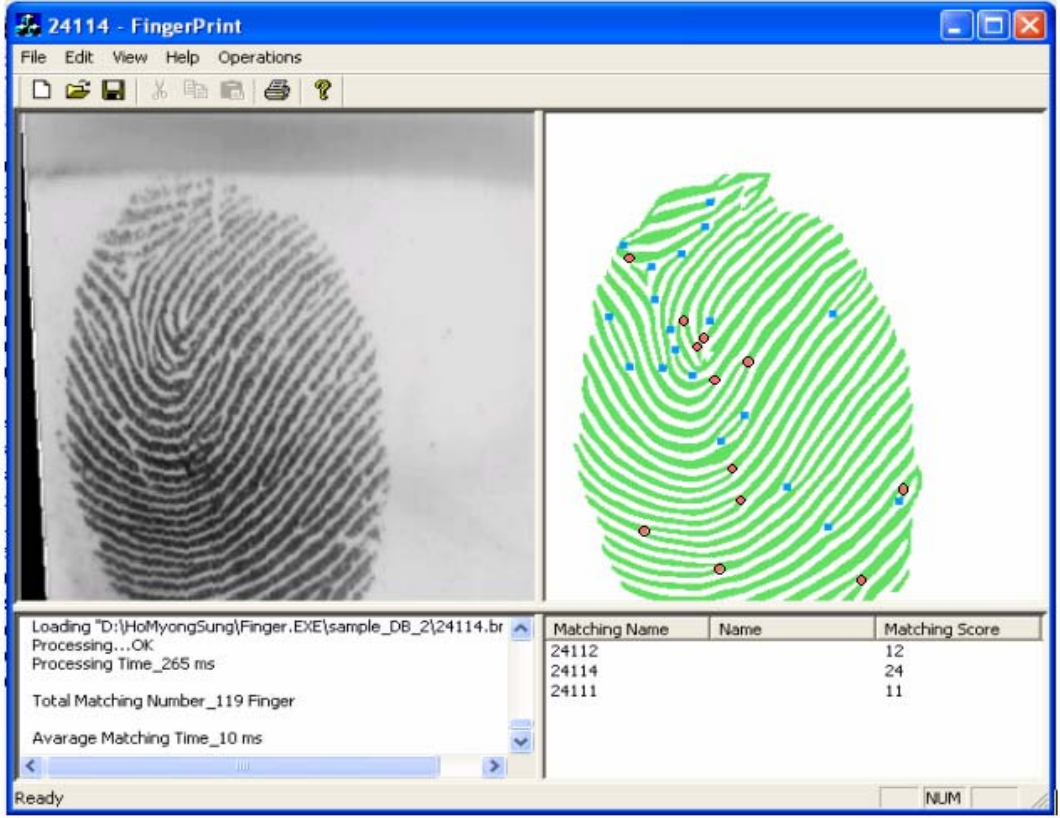

Figure. Off-line Fingerprint verification